\pdfoutput=1

\documentclass[10pt,journal,compsoc]{IEEEtran}
\usepackage[pdftex]{graphicx} 
\usepackage[export]{adjustbox}
\usepackage{url} 
\usepackage{amsmath}
\usepackage{amssymb}
\usepackage{bm}
\usepackage{booktabs,siunitx,caption}
\usepackage{subcaption}
\usepackage{graphicx}
\usepackage{float}
\usepackage{booktabs} 
\usepackage[hidelinks]{hyperref} 
\usepackage{multirow}
\usepackage{multicol}
\usepackage{array}
\usepackage{booktabs}
\usepackage{tikz}
\usepackage[normalem]{ulem}

\ifCLASSOPTIONcompsoc
  \usepackage[nocompress]{cite}
\else
  \usepackage{cite}
\fi
\ifCLASSINFOpdf
\else
\fi
\begin{document}
%
\title{Self-Supervised Approach for\\ Facial Movement Based Optical Flow}
%
%
%
%

\author{Muhannad~Alkaddour,
        Usman~Tariq,~\IEEEmembership{Member,~IEEE,}
        and~Abhinav~Dhall,~\IEEEmembership{Member,~IEEE}
\IEEEcompsocitemizethanks{\IEEEcompsocthanksitem M. Alkaddour was with the Graduate Program in Mechatronics Engineering, American University of Sharjah, Sharjah,
UAE.\protect\\
E-mail: malkaddour@aus.edu
\IEEEcompsocthanksitem U. Tariq is with the Department of Electrical Engineering, American University of Sharjah, Sharjah, UAE.\protect\\
E-mail: utariq@aus.edu
\IEEEcompsocthanksitem A. Dhall is with the Department of Human Centred Computing, Monash University, Melbourne, Australia, and the Department of Computer Science and Engineering, Indian Institute of Technology Ropar, Rupnagar, India.\protect\\
E-mail: abhinav.dhall@monash.edu}
\thanks{Manuscript received October X, 2020; revised XXXXX XX, 202X.}}

%
%

\markboth{IEEE Transactions on Affective Computing,~Vol.~11, No.~13, July-September~2020}%
{Alkaddour \MakeLowercase{\textit{et al.}}: Self-Supervised Approach forFacial Movement Based Optical Flow}
%



\IEEEtitleabstractindextext{%
\begin{abstract}
Computing optical flow is a fundamental problem in computer vision. However, deep learning-based optical flow techniques do not perform well for non-rigid movements such as those found in faces, primarily due to lack of the training data representing the fine facial motion. We hypothesize that learning optical flow on face motion data will improve the quality of predicted flow on faces. The aim of this work is threefold: (1) exploring self-supervised techniques to generate optical flow ground truth for face images; (2) computing baseline results on the effects of using face data to train Convolutional Neural Networks (CNN) for predicting optical flow; and (3) using the learned optical flow in micro-expression recognition to demonstrate its effectiveness. We generate optical flow ground truth using facial key-points in the BP4D-Spontaneous dataset. The generated optical flow is used to train the FlowNetS architecture to test its performance on the generated dataset. The performance of FlowNetS trained on face images surpassed that of other optical flow CNN architectures, demonstrating its usefulness. Our optical flow features are further compared with other methods using the STSTNet micro-expression classifier, and the results indicate that the optical flow obtained using this work has promising applications in facial expression analysis.
\end{abstract}

}

\maketitle

\IEEEdisplaynontitleabstractindextext

%
\IEEEpeerreviewmaketitle

\IEEEraisesectionheading{\section{Introduction}\label{sec:introduction}}

\IEEEPARstart{F}{acial} expressions are generated due to non-rigid movement in faces. From the perspective of automatic facial expression recognition (FER), the motion information has been well explored for the task of both micro and macro expression analysis. Optical flow is used to estimate the motion of sets of pixels across images. This information on faces can help characterize both micro and macro expressions, which are useful in expression recognition. A major motivation for using the motion information for FER is based on what is known as the facial feedback hypothesis \cite{facial_feedback}, which, in summary, suggests that facial actions can both encode current emotions as well as \textit{induce} or amplify emotions. An example of this would be that the furrowing of the brow could increase anger \cite{facial_feedback}. It has also been demonstrated that some facial muscle movements are linked to the compound facial expression of negation \cite{not_face}. Also, the relation between motion information extracted from the eyes and mouth has been studied in its association with the facial expressions of psychopaths \cite{psychopath}. Facial and head movements are also important in social contexts, such as head motion used to indicate particular social cues, or the famous twitching of the lip corners that may suggest lying \cite{handbook}.

Faces have a peculiar structure. Hence, in this work, we focus on learning optical flow specialized for faces which we will attempt to constrain the algorithm to learn only lifelike expressions on faces. In doing so, we explore how well a deep network can perform in this task. We demonstrate that the proposed architecture will work well for faces compared to traditional optical flow algorithms. The results can serve as a precursor to designing motion-based features for supervised and unsupervised learning of facial expressions by drawing on existing research linking facial motion information to facial expression and emotion recognition. Several works document the use of facial optical flow features for facial expression recognition and action unit recognition tasks.

We use the BP4D-Spontaneous dataset \cite{bp4d} consisting of videos of 41 participants with different facial expressions to generate the ground-truth optical flow between every pair of consecutive frames in the dataset. The ground-truth optical flow is obtained using facial key-points and image warping with affine transformations. We then use this facial optical flow ground truth to train a convolutional autoencoder based architecture, FlowNetS \cite{flownet} (specialized for optical flow estimation), to learn optical flow specialized for facial motions, meaning that the motion learned should exhibit local coherency as would be expected on faces. We also modify the architecture by adding a cyclic loss to help the network reconstruct the latter image in a given image pair using the optical flow predicted by the network. We argue that adding this reconstruction in the learning framework improves the predicted optical flow by guiding it using the structure of the image pairs. We perform an ablation study with different loss functions, and compare the performance of our network and other baseline optical flow CNNs. Finally, we test the usefulness of our network by using the learned optical flow predictions for micro-expression detection using optical flow and the Shallow Triple Stream Three-dimensional CNN (STSTNet) \cite{optical_flow_cnn_3}.

Hence, the contributions of this paper are: \begin{itemize}
    \item Introduction of a \emph{``noisy''} optical flow dataset for faces, making use of the peculiar structure of faces.
    \item Learning a network for optical flow estimation, specialized for face movements. We then complement the structure with a cyclic loss. Our modified architecture outperforms several other networks used for optical flow estimation.
    \item Exhibiting the usefulness of our trained network by applying it for micro-expression detection.
\end{itemize}

The remainder of the paper is organized as follows. Section \ref{sec:related} contains related literature in the relevant topics.
Section \ref{sec:dataset} describes the details of the automatic dataset generation used in this paper, and details of the networks trained on the generated dataset are explained in Section \ref{sec:baseline}. The results of the ablation study and micro-expression recognition are presented in Section \ref{sec:ablation}. And finally, we  present the concluding remarks and recommendations for improvement and future work in Section \ref{sec:conclusion}.

\section{Related Work}\label{sec:related}
First, we discuss works related to optical flow estimation using classical and deep learning techniques, along with some of the common challenges. We follow this up by a survey of optical flow methods as applied to faces in particular, and how optical flow is used in tasks such as micro-expression detection.
\subsection{Optical Flow Estimation}\label{subsec:optical_flow}
Optical flow in images is used to estimate the motion of sets of pixels across images. 
Classical methods, such as in \cite{prop20} and \cite{bahar}, use the intensity derivatives and energy methods to estimate the optical flow. However, there may be several challenges.

\subsubsection{Optical flow challenges} \label{subsec:optical_flow_challenges}
Different methods have been proposed to compensate for typical problems that may arise in optical flow estimation. Chen \textit{et al.} \cite{prop25} developed a method using quaternions to deal with the possible inconsistency among the RGB channel intensities, Portz \textit{et al.} \cite{prop26} proposed an algorithm to compute optical flow in blurred environments, and Porikli \textit{et al.} \cite{prop27} modified the optical flow algorithm to deal particularly with low frame-rate applications. Finally, Zappella \textit{et al.} \cite{prop28} presented a comprehensive literature review and evaluation of motion tracking algorithms, including their advantages and applications. 
\subsubsection{Deep learning for optical flow estimation}
\label{subsubsec:optical_flow_deep_learning}
With the surge and success of deep learning applications this decade, there has also been a rise in using convolutional neural networks to learn optical flow, beginning with the groundbreaking work of Fischer \textit{et al.} \cite{flownet} with their \textit{FlowNet} CNN architecture. Building on the success of FlowNet, \textit{FlowNet2.0} \cite{flownet2} was introduced a few years later to improve performance by stacking networks, scheduling the training data, and learning small-motion datasets. \textit{FlowNet3.0} \cite{flownet3} was also proposed afterwards for scene flow estimation. For our experiments, we use the \textit{FlowNetS} architecture adapted from \cite{flownet} to train on our dataset. By demonstrating how we can adapt FlowNetS to perform well on datasets consisting of only faces, we can later improve even further by training the data with the more advanced architectures.

While FlowNet is one of the most popular optical flow deep learning architectures, several other architectures have since been proposed to deal with certain challenges. Janai \textit{et al.} \cite{unsupervised_learning_good} dealt with the problem of unsupervised learning of optical flow in occluded settings by considering a triplet instead of a pair of frames and a \emph{photometric} loss to handle the occlusions. Ren \textit{et al.} \cite{unsupervised_flow} and Meister \textit{et al.} \cite{unflow} also built on these concepts for unsupervised learning of optical flow. 
Sun \textit{et al.} \cite{optical_pyramid} used the pyramid-structure CNN architecture \textit{PWC-Net} for optical flow prediction, which we use in this work to test on the face optical flow dataset as a benchmark implementation and compare with our performance. Another optical flow CNN we use for comparison in this work is \textit{LiteFlowNet} by Hui \textit{et al.} \cite{liteflownet}, which surpassed Flownet2.0's performance on the KITTI and Sintel final datasets.

In their pioneering work, Zhu \textit{et al.} \cite{cycle_gan} developed the \textit{cycleGAN}, which is a type of generative adversarial network (GAN), that implements a cyclic loss function which is used as a metric to evaluate the network's prediction as compared with one of the inputs. This loss function is also used in the context of optical flow learning. Yu \textit{et al.} \cite{cyclic_faces} used this cyclic loss, which they dub \emph{``warp loss''},  to train a Flownet architecture for optical flow learning. The cyclic loss is also adapted by Lai \textit{et al.} \cite{cyclic_faces_semi} in the context of optical flow using a GAN. Both of the latter architectures used a differentiable spatial transformer layer with learnable parameters, adapted from Jaderberg \textit{et al.} \cite{spatial_transformer}.

\subsection{Optical Flow and Facial Expression Analysis} \label{subsec:face_optical_flow} 
In this section, we discuss various convolutional neural network architectures for optical flow estimation as well as deep learning that uses optical flow for facial expression analysis.
\subsubsection{Face optical flow estimation}
More relevant to our topic are motion tracking methods, which are used in facial expression analysis. One important work in learning optical flow for facial expressions by Snape \textit{et al.} is \textit{Face Flow} \cite{faceflow}, which minimizes a proposed energy to learn the flow field for a sequence of frames consisting of facial expressions. Another relevant work is optical flow dataset generation done by Le \textit{et al.} \cite{unsupervised_generation} who are also concerned with producing optical ground-truth data for general video sequences. According to them, little prior work exists on how the performance of CNNs is influenced by optical flow datasets, and their main focus is that of non-rigid motion. Our work can be considered to be a contribution to the study of optical flow's effects on CNNs, with the difference being that we focus on facial datasets instead. We attempt to learn optical flow from the face movements themselves. On a side note, a review of different optical flow techniques specialized for facial expression recognition can be found in \cite{opt_flow_techniques}.

\subsubsection{Face optical flow and deep learning}
\label{subsubsec:face_optical_flow_deep_learning}
We mention a few implementations of deep learning in facial expression analysis using optical flow. Koujan \textit{et al.} \cite{deepfaceflow} recently proposed \textit{DeepFaceFlow}, in which they construct a 3D optical flow dataset for faces from a large collection of videos and compare the performance of their U-net trained on their dataset with other CNN architectures for both 2D and 3D optical flow estimation. One key difference between our work and theirs is that we incorporate a cyclic loss to test how well the flow field reconstructs the second image in the pair. Additionally, the training data we generate is based upon the BP4D-Spontaneous dataset, which is specifically tuned to exhibit various emotions and thus more specialized for expression recognition tasks. We also test our network's performance on microexpression detection.

Several works also use optical flow for action unit recognition. Ma \textit{et al.} \cite{AUR} proposed Action Unit (AU) R-CNN to improve AU recognition by using expert prior knowledge, which can be in the form of optical flow, to guide an R-CNN in locating the action region. Yang and Yin \cite{AUR_1} learn both optical flow and facial action units for static images in one combined CNN architecture. 
Other works that use optical flow for action unit recognition can be found in \cite{AUR_2}, \cite{AUR_3}, and \cite{AUR_4}. 

Liong \textit{et al.} \cite{apexNet} exploit the optical flow in a video sequence between the frame with the highest intensity, called apex, and each of the rest of the frames, using the optical flow as input to a deep network for micro-expression detection.
They also use apex and onset frames in \cite{optical_flow_cnn_3} to compute optical flow along with an added feature, the optical strain, as input to STSTNet, which we adapt in this work to test for micro-expression recognition.
Verburg and Menkovski \cite{optical_flow_cnn_2} use optical flow histograms as feature inputs to a recurrent neural network for the recognition of micro-expressions. Li \textit{et al.} \cite{fusing} use a CNN to locate facial keypoints and FlowNet2.0 to compute optical flow, and the flow features are then used with a support vector machine for micro-expression detection. 


\section{Dataset Preparation}\label{sec:dataset}
Our method is inspired by the progress in self supervised learning techniques for action recognition \cite{wang2015unsupervised} and eye gaze prediction \cite{dubey2019unsupervised}. We use the BP4D-Spontaneous dataset \cite{bp4d}, which consists of 41 subjects with 8 video sequences each, containing videos of elicited emotions. 
The motivation for using BP4D-Spontaneous is its inclusion of both head and facial motion. While local non-rigid facial motion estimation is the primary focus of this work, it is also useful to capture this local facial flow in the presence of head motion. Since BP4D-Spontaneous is concerned with spontaneously elicited expression sequences and 3D encoding, more general motion is available. Other datasets, such as the Extended CK+ \cite{CK}, are more specialized for AU or micro-expression detection, and thereby are less suited for a more general motion framework. Moreover, this allows us to test how optical flow performs on micro-expression detection when trained on a dataset not specialized for micro-expression detection.

\begin{figure*}[t]
    \centering
    \includegraphics[trim=54 56 78 20,clip,width=\linewidth]{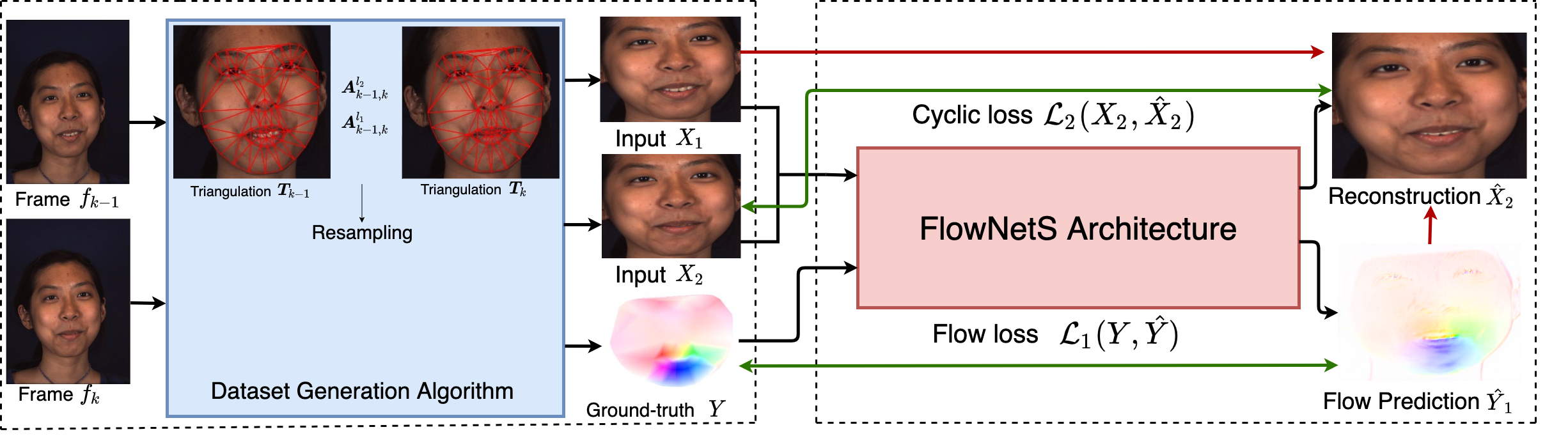}  
    \vspace{-4mm}
    \caption{Overall pipeline for data generation and network training: Two examples of the affine maps are shown for some triangles $l_1$, $l_2$, and an illustration of the resampling process is shown on a $3\times3$ grid of a portion of the optical flow field.}  \vspace{-2mm}
    \label{fig:flowchart}
\end{figure*}

Fig. \ref{fig:flowchart} shows the overall pipeline for a pair of frames and how they can be used for dataset generation and CNN training.\footnote{Our code implementing the algorithm in this section will be made publicly available.}
We introduce the notation that we'll use throughout this section to generate the optical flow ground truth from the BP4D-Spontaneous dataset \cite{bp4d}. 
For a given sequence $S$ in the dataset, we denote the frames contained in $S$ by $F=\{f_k\}_{k=0}^{N_f}$, where $f_{k}\in \mathbb{R}^{H\times W \times 3}$ are the ordered frames. Our aim in this section is to compute a set of optical flow fields, $\boldsymbol{U}$ separately for every ordered set of frames, $F$, where $\boldsymbol{U}=\{\boldsymbol{u}_k\}_{k=0}^{N_f-1}$ contains the optical flow fields $\boldsymbol{u}_k: \mathbb{R}^{H\times W} \mapsto \mathbb{R}^{H\times W \times 2}$ for each frame except the final one in that sequence. The $\boldsymbol{u}_k$ are vector-valued functions defined on the image grid. 

Landmarks $\boldsymbol{P}$ on the face in $S$ are tracked for each frame using the open source OpenFace pipeline \cite{openface2}, which uses the \textit{Convolutional Experts Constrained Local Model} \cite{openface1} to obtain $68$ landmarks per face. We note that, for this step, other facial keypoint detection techniques can also be used. We denote the facial landmarks tracked on the face in \textit{each} frame $f_k$ of $S$ by $\boldsymbol{P}_k=\left(\boldsymbol{p_0}\\ \hdots \\\boldsymbol{p_{68}} \right)_{k}^T \in \mathbb{R}^{68\times 2}$  (where, $T$ denotes the transpose operation).

Next, we completely partition the first face $f_0$ into a triangular mesh using Delaunay triangulation on $\boldsymbol{P}_0$ using Scipy's Delaunay triangulation package. Theoretical background related to Delaunay triangulation can be found in \cite{affine_book}.
This mesh divides the face in $f_0$ into $N_t$ disjoint triangles $\boldsymbol{T}_0=\{\boldsymbol{t}^l\}_{l=0}^{N_t}$, where each $\boldsymbol{t}^l = \begin{pmatrix}\boldsymbol{v}_0, & \boldsymbol{v}_1, & \boldsymbol{v}_2\end{pmatrix}_l^T \in \mathbb{R}^{3\times 2}$ is the matrix with rows composed of vertices of triangle $l$. After triangulating $f_0$, we use similar triangulation on the remaining frames in the sequence, yielding the set of triangulations $\{\boldsymbol{T}_k\}_{k=0}^{N_f}$ on $S$. 

We use the triangulation $\boldsymbol{T}_{k-1}$ to capture the local motion on every triangle in the face partition from frame $f_{k-1}$ to frame $f_{k}$. 
Given the triangle $\boldsymbol{t}_{k-1}^l$, we infer an affine map $\boldsymbol{A}_{k-1}^l \in \mathbb{R}^{3\times3}$ that sends its vertices to the vertices in $\boldsymbol{t}_{k}^l$. Specifying three mappings are sufficient to uniquely define an affine map \cite{geometric_methods}. We can define $\boldsymbol{t}^*=\begin{pmatrix}\boldsymbol{t}, & \boldsymbol{1}_{3\times 1}\end{pmatrix}^T\in \mathbb{R}^{3\times3}$ to be the matrix of homogeneous coordinates of each vertex. Then, for all triangles in $f_{k-1}$ and $f_k$, $\boldsymbol{A}_{k-1}^l$ that sends $\boldsymbol{t}_{k-1}^*$ to $\boldsymbol{t}_k^*$ is uniquely determined by,
\begin{align}
    \boldsymbol{A} = \left(\boldsymbol{t}_{k-1}^*\right)^{-1}\boldsymbol{t}^*_k.
\end{align}
This gives the required matrix for the affine map. Note that if the triangle is degenerate, then $\boldsymbol{t}_{k-1}^*$ will be singular.
Once the correspondence between the two triangles across frames is known, $\boldsymbol{A}_{k-1}^l$ also maps the interior of $\boldsymbol{t}_{k-1}^l$ to the interior of $\boldsymbol{t}_{k}^l$, since barycentric coordinates are invariant under affine maps \cite{geometric_methods}.

We use the barycentric coordinates to compute the interiors of all the triangles in $\boldsymbol{T}_0$, and then learn each affine map $\boldsymbol{A}_0^l$ as described above to map all the triangle interiors from $\boldsymbol{T}_0$ to $\boldsymbol{T}_1$. To compute the interior of the triangle using barycentric coordinates, an efficient algorithm from \cite{collision-detection} can be used to test if an arbitrary point $\boldsymbol{v}$ is contained in a given triangle by taking the convex combination with the triangle vertices
\begin{equation}
\label{eq:barycentric_triangle}
\begin{aligned}
    \boldsymbol{v} = (1-\lambda_1 - \lambda_2)\boldsymbol{v}_0 + \lambda_1 \boldsymbol{v}_1 + \lambda_2 &\boldsymbol{v}_2 \\
    \boldsymbol{v} - \boldsymbol{v}_0 =  \lambda_1 (\boldsymbol{v}_1 - \boldsymbol{v}_0) + \lambda_2 (\boldsymbol{v}_2 - &\boldsymbol{v}_0).
    \end{aligned}
\end{equation}
By taking the dot product of equation (\ref{eq:barycentric_triangle}) with $\boldsymbol{v}_1 - \boldsymbol{v}_0$ and $\boldsymbol{v}_2 - \boldsymbol{v}_0$, a $2\times2$ system of equations can be solved for $\bm{\lambda}=\begin{pmatrix}\lambda_1, & \lambda_2\end{pmatrix}^T$, and $\lambda_3 = 1 - \lambda_1 - \lambda_2$ \cite{collision-detection},
\begin{equation}
\begin{gathered}
\label{eq:barycentric_solution}
    \bm{V}\bm{\lambda} = \bm{b} \text{, where,} \\
    \bm{V}= \begin{pmatrix} \lVert \boldsymbol{v}_1 - \boldsymbol{v}_0 \rVert_2^2 & (\boldsymbol{v}_2 - \boldsymbol{v}_0) \cdot (\boldsymbol{v}_1 - \boldsymbol{v}_0) \\ (\boldsymbol{v}_2 - \boldsymbol{v}_0)\cdot(\boldsymbol{v}_1 - \boldsymbol{v}_0) & \lVert \boldsymbol{v}_2 - \boldsymbol{v}_0 \rVert_2^2 \end{pmatrix} \text{, and}\\
    \bm{b}=\begin{pmatrix}(\boldsymbol{v} - \boldsymbol{v}_0)\cdot (\boldsymbol{v}_1 - \boldsymbol{v}_0) \\ (\boldsymbol{v} - \boldsymbol{v}_0)\cdot (\boldsymbol{v}_2 - \boldsymbol{v}_0)\end{pmatrix}
\end{gathered}
\end{equation}
and if the $\lambda_i \in \left[0, 1\right]$, then $\boldsymbol{v}$ lies in the closure of the triangle of interest, i.e. $\boldsymbol{v}$ is a convex combination of the columns of $\boldsymbol{t}$. We test all points in this way using a rectangular discrete grid surrounding the triangle. Repeating this for all $f_k$ in the sequence is overall computationally expensive, so we only do it for triangles in the first frame of that video. By invariance of barycentric coordinates under the affine maps $\boldsymbol{A}_{k-1}^l$, this also determines the barycentric coordinates for all subsequent frames $f_k$, $k > 0$.

After determining the affine maps and mapping the triangles and their interior pixels $\boldsymbol{v}_{k-1}$ to $\boldsymbol{v}_k$, we compute the per-pixel optical flow vector $\Tilde{\boldsymbol{u}}_{k-1}$ by
\begin{equation}
\label{eq:optical_flow_u}
    \Tilde{\boldsymbol{u}}_{k-1} = \boldsymbol{v}_k - \boldsymbol{v}_{k-1}.
\end{equation}
However, it is not guaranteed that the domain of any given affine map, $\boldsymbol{A}$, will lie on an discrete grid $\mathbb{Z}^+ \times \mathbb{Z}^+$. When the domain is not a discrete grid, the optical flow fields $\Tilde{\boldsymbol{u}}_k$ are defined on points that are not necessarily pixel coordinates, which affects the frames \textit{after} $f_0$. The optical flow field $\Tilde{\boldsymbol{u}}_{k-1}$ from equation (\ref{eq:optical_flow_u}) is defined on a discrete grid, but the pixels that are mapped from $f_0$ to $f_1$ will subsequently be mapped from $f_1$ to $f_2$, in which case it is not guaranteed that they also lie on a discrete grid. To recover the optical flow field $\boldsymbol{u}_{k-1}$ on a discrete grid in the target image, we use bicubic spline interpolation over the irregular grid using $\Tilde{\boldsymbol{u}}_{k-1}$. We only do this to define the optical flow field at each frame, but continue to learn the affine maps on the irregular grids, since we wish to preserve the same barycentric coordinates obtained in $f_0$ for all frames. The flow fields are stored in .flo formats for later use in the experiments. 

Together with the resampling stage, this procedure gives us the ground-truth vector field for all pixels of frame $f_{k-1}$. The details can be summarized as follows:
\begin{enumerate}
    \item Starting from frame $f_0$, determine the interiors of all triangles $\boldsymbol{t}_0^l$, using barycentric coordinates.
    \item Learn the affine maps sending all $\boldsymbol{t}_{0}^l$ to $\boldsymbol{t}_1^l$ and transform the entire face to obtain the first optical flow field $\boldsymbol{u}_0$.
    \item For all frames starting from $f_1$, again infer the affine maps sending all $\boldsymbol{t}_1^l$ to $\boldsymbol{t}_2^l$ and apply the transformation on all the pixels which have already been mapped from frame $f_{0}$. This removes the need to expensively compute the triangle interiors for frame $f_1$ while still finding the optical flow field $\Tilde{\boldsymbol{u}}_1$.
    \item From $\Tilde{\boldsymbol{u}}_1$, resample the flow field over a discrete grid to yield the ground-truth flow $\boldsymbol{u}_1$.
    \item Repeat steps 3-4 for the remaining frames in a sequence $\{f_k\}_{k=2}^{N_f}$, for all sequences and subjects.
\end{enumerate}The total number of images in the generated dataset is 325720, and these were partitioned into 228171, 65130, and 32419 for training, validation, and test data respectively. The dataset generation was completed in a total of about five days using multicore CPU parallel processing with four parallel processes running at a time.
\section{Baseline Networks}\label{sec:baseline}
In this section, we describe the CNN architecture used to train the optical flow, followed by the training and ablation study details. These details include the different hyperparameters used in the different experimental setups, such as the choices of loss functions, the loss weights, and training/testing data split. 
\subsection{CNN architecture: \textit{FlowNetS}}\label{sec:FlowNet}
To test the effects of having a large, \emph{``noisy''} ground-truth optical flow dataset specialized for faces on CNNs, the FlowNetS \cite{flownet} architecture was used. FlowNetS is one of the pioneering CNNs on optical flow learning. While more sophisticated optical flow architectures have been developed, our purpose is to demonstrate the improvement of training a CNN with face data compared with some other datasets, e.g., the FlyingChairs dataset, as a proof of concept. Should we discover an improvement, in future, we can expand it to tackle other problems (e.g. robustness to occlusion).

FlowNetS is a convolutional autoencoder architecture which accepts a pair of images as input and outputs the per-pixel optical flow from the first image to the second. It consists of a sequence of downsampling convolutional layers in the encoder followed by upsampling layers in the decoder, in addition to intermediate operations and concatenations.
Another variant of FlowNet, which is FlowNetCorr, is characterized by a cross-correlation layer which fuses two input streams together, contrasted to FlowNetS which combines them with a simple concatenation. The difference in performance reported in \cite{flownet} is not too significant, and including the cross-correlation layer during training resulted in the inconvenience of much longer training times.
The output resolutions of each of the flow predictions in our network are slightly different than the original FlowNetS. Specifically, the ratio of our flow prediction heights to theirs is 24:17, and our widths to theirs is 4:5. The reader is referred to \cite{flownet} for specific details on the network architecture. 
\subsection{Cyclic loss for image reconstruction}
\label{subsec:cyclic_loss}
\begin{figure*}[t]
 	\setlength{\fboxsep}{0pt}%
 	\setlength{\fboxrule}{0pt}%
 	\centering
 	    \includegraphics[trim=0 0 0 22,clip,width=\linewidth]{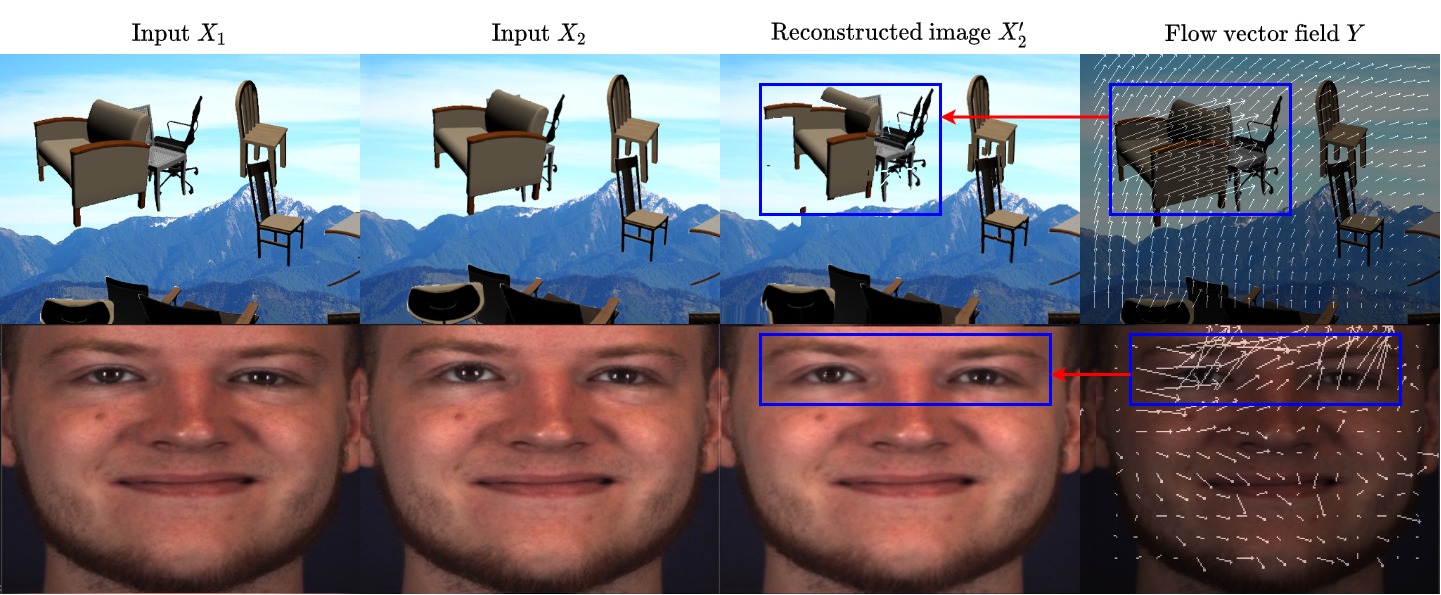}  
    \vspace{-6mm}
 	\caption{Effect of using flow field $Y$ to warp $X_1$ to $X_2^\prime$ is demonstrated for images with large (top) and small (bottom) motion.\label{fig:warped_examples}}
 \end{figure*}

For some of the experiments described in the next section, a cyclic loss is implemented to minimize the difference between the output predicted using the flow prediction and the second input image. This resulted in an additional warping layer to the network that acts on the flow prediction with highest resolution. The warping layer uses the predicted per-pixel flow field vectors to warp the first input image, and the result is recovered using bilinear interpolation. We note that structures inherent only to the second input cannot be reproduced in the warped output, since the warping function only changes pixel locations from the first input, and does not contain any learnable parameters. Fig. \ref{fig:warped_examples} shows two examples of this phenomena from FlyingChairs and our face dataset, showing the original input image pair $(X_1$, $X_2)$, the image $X_2^\prime$ deformed using the flow field, and visualization of the flow field $Y$.

The dominant motion in the FlyingChairs image pair from the flow field is rightward motion of the left armchair. The location of the armchair in the warped image is correct, but the reconstruction of the warped portion is missing. This is also present in the smaller desk chair, making a copy of itself at the warped location during reconstruction. Due to these large differences in the images, adding a warping layer while training on the FlyingChairs dataset is likely to worsen the network's performance. However, this effect is much more subtle in our face dataset due to the higher frame rate of the sequences, which causes lower magnitude motion between every two consectuive frames.
For the face example in Fig. \ref{fig:warped_examples}, the deformed image $X_2^\prime$ is perceptually similar to the actual $X_2$, particularly in the upwards motion of the eyes and the slight rightward motion caused by the furrowing of the brow. Since the time difference between two frames is very small in the face dataset, it is very unlikely for new structures to be introduced in $X_2$. A notable exception to this is the opening (closing) of the mouth due to revealing (hiding) teeth, which cannot be reproduced by pixel rearrangement alone. Another exception would be the squinting or widening of the eyes for the same reason, since the eyelid or eyeball would not be present in the first image. Although the artifacts caused by the warping produced a flawed image in the FlyingChairs dataset, we hypothesize and show that it still helps guide the directions of the predicted flow when training on faces since the undesirable effects are considerably less due to the lower amount of new structure.

\subsection{Training and ablation studies details}\label{subsec:training_details} 
The training details of the aforementioned architecture are described in this section\footnote{Code for the details described in this section will be made publicly available.}. Ablation studies are performed on FlowNetS by training the network with different loss functions and their corresponding weights. 

We denote by $(X_i, X_{i+1})$ the pair of successive input frames, where $X_i, X_{i+1} \in \mathbb{R}^{384\times512\times3}$, $Y_i \in \mathbb{R}^{384\times512\times2}$ is the ground-truth flow field, and $\hat{\boldsymbol{Y}}_i = \{(\hat{Y}_i)_k\}_{k=1}^5$ contains the intermediate multi-scale flow field predictions, where each element $(\hat{Y}_i)_k\in \mathbb{R}^{H_k\times W_k\times2}$. The $i$ enumerates the entire training set, and successive image frames are input to the network at every iteration. The resolutions of the flow predictions are $(H_k, W_k)=(384\times2^{-k}, 512\times2^{-k})$ for $k\in\{1,\ldots, 5\}$ in the decoder. $(\hat{Y}_i)_1$ is the largest flow prediction, as in the original FlowNetS output. 
Note that, in the following, we drop the added subscript and refer to it as $\hat{Y}_i$.

Since we assume that the background is stationary, much of the ground-truth flow field outside of the boundaries defined by the key-points are zero vectors. To make the training more practical, we zoom on the box with vertices defined by the key-points with maximal and minimal coordinates plus some offset in the $x$ and $y$ directions. The cropped images and flow fields are then resized using bilinear interpolation. To preserve the units of the flow vectors as pixels, they are scaled accordingly in the horizontal and vertical directions. 

Next, we describe the different experimental setups used to train the networks.

\subsubsection{Experimental setup 1: no cyclic loss}\label{subsubsec:experiment_1}
In this experiment, the architecture is used without the additional warping layer. The network was trained for 30, 40, and 400 epochs on the face, FlyingChairs, and Sintel datasets respectively, with 15000, 21592, and 870 training and 1000, 640, and 271 validation input image pairs each. The batch size used for training is 16 input pairs. The loss function is the average endpoint error (EPE), $\mathcal{L}_1^i(Y_i, \hat{\boldsymbol{Y}}_i)$, defined for one output by,
\begin{equation}
\label{eq:EPE}
    \mathcal{L}_1^i(Y_i, \hat{\boldsymbol{Y}}_i) = \sum_{k=1}^5 \frac{w_k}{H_kW_k}\sum_{j=1}^{H_kW_k} \left\lVert{\boldsymbol{y}_{ij}-(\hat{\boldsymbol{y}}_{ij})_k}\right\rVert_{2}. 
\end{equation}

Here, the $w_k$ are loss weights for each intermediate flow prediction loss, given by $w_k = 2^{-k}$. $H_k, W_k$ are the sizes of the intermediate predictions and the $\boldsymbol{y}_{ij}, (\hat{\boldsymbol{y}}_{ij})_k$ are the flow vectors for the $j$th pixel of ground-truth and $k$th predicted flow fields $Y_i$ and $(\hat{Y}_i)_k$. The flow fields $Y_i$ are resized to compute the error for each intermediate prediction. The optimizer used is Adam, with $\beta_1=0.9$ and $\beta_2=0.999$ as in \cite{flownet}. This performs better than alternative optimizers. We initialized the learning rate $\alpha$ at $1e{-4}$ for faces and $5e{-5}$ for FlyingChairs and scheduled similar to \cite{flownet}. 

For preliminary experimentation, we trained the network once on the face data for a $15$k and $1$k training and testing split, and once separately on the entire FlyingChairs dataset. We then tested each trained network on both datasets each, as well as the Sintel dataset \cite{sintel}. After the preliminary experiment, we trained the same network again from scratch on faces only for a $228$k, $65$k, and $32.5$k train/val/test split, exactly as in the next two experiments, to make them comparable. The latter setup is referred to as \emph{Experiment 1}, from here onwards.

\subsubsection{Experimental setup 2: with cyclic loss}\label{subsubsec:experiment_2} 
When the warping layer \cite{cyclic_faces_semi} at the end of the network is included, it is necessary to define a cyclic loss function for the warped output $\hat{X}_{i+1}$ and the second input $X_{i+1}$. We expect to see an improvement in the flow prediction due to the cyclic loss. For this experiment, we define the additional cyclic loss function $\mathcal{L}_2^i(X_{i+1}, \hat{X}_{i+1})$ for one output pair $i$ as:
\begin{equation}
\label{eq:cyclic_loss}
\begin{aligned}
    \mathcal{L}_2^i(X_{i+1}, \hat{X}_{i+1}) = \frac{1}{HW}\sum_{j=1}^{HW}\frac{1}{3}\sum_{k=1}^{3}\left\lVert x_{i+1,j,k} - \hat{x}_{i+1,j,k}\right\rVert_{H_1} \\
    \left\lVert x \right\rVert_{H_1} =  \begin{cases} 
      \frac{1}{2}x^2 &  |x|\leq d \\
      \frac{1}{2}d^2 + d(|x|-d) & |x|> d
   \end{cases}
\end{aligned}
\end{equation}
which uses the Huber loss function $\lVert x \rVert_{H_1}$\cite{huber1964}, a variant of the $L_1$ loss that is everywhere differentiable, since it is quadratic for small values of $x$. The $x_{i+1,j,k},\, \hat{x}_{i+1,j,k}$ are values of the $j$th pixel of $X_{i+1},\,\hat{X}_{i+1}$ at color channel $k$. 
We also note that $\hat{X}_{i+1}$ is a function of the first image of the input pair, $X_i$, and $\hat{Y}_i$, which is the flow prediction with largest resolution. 

The total loss function $J(X, \hat{X}, Y,\hat{\boldsymbol{Y}})$ for all training pairs is then 
\begin{equation}
    \label{eq:total_cyclic}
    J(X, \hat{X}, Y,\hat{\boldsymbol{Y}}) = \frac{1}{M}\sum_{i=0}^{M-1}\left[\lambda_1\mathcal{L}_1^i(Y_i,\hat{\bm{Y}}_i) + \lambda_2\mathcal{L}_2^i(X_{i+1}, \hat{X}_{i+1})\right]
\end{equation}
with the $\mathcal{L}_1^i, \mathcal{L}_2^i$ defined in equations (\ref{eq:EPE}) and (\ref{eq:total_cyclic}) and $\lambda_1, \lambda_2$ to be specified, averaged over all $M$ training examples.
In this experiment, we train the network on both faces and FlyingChairs datasets using two different sets of loss weights $\lambda_1,\ \lambda_2$. One network has more emphasis on reconstruction, with $\lambda_2=0.6,\ \lambda_1=0.4$. We refer to this as Case I. The other network has higher weight assigned to the EPE with $\lambda_1=0.75,\ \lambda_2=0.25$. We refer to this as Case II. Note that the $w_i$ in equation (\ref{eq:EPE}) should sum to $\lambda_1$. 
For both cases, we trained the network on faces for 15 epochs and 228160 training pairs. Learning rates were kept constant for these experiments throughout training, since scheduling them as previously done lead to very large gradients halfway through training. In Case I, the learning rates were $2.5e{-6}$ and $1.25e{-6}$ for faces and FlyingChairs respectively, and in Case II, they were both set to $2.5e{-6}$. We then tested the trained networks on the test set of 32416 image pairs.

\subsubsection{Experimental setup 3: with cyclic loss, smoothness constraint, and average angular error}\label{subsubsec:experiment_3} 
In this experiment, we added an additional loss function $\mathcal{L}_3^i(\hat{Y}_i)$. In Case I of this experiment, a smoothness constraint was imposed on the flow prediction by minimizing the flow gradients, defined as:
\begin{equation}
\begin{aligned}
    \label{eq:smoothness_constraint}
    \mathcal{L}_3^i(\hat{Y}_i) = \frac{1}{HW}\sum_{j=1}^{HW} \Bigg(&\left\lVert \frac{\partial\hat{u}_{ij}}{\partial x} \right\rVert_{H_1}+\left\lVert \frac{\partial\hat{u}_{ij}}{\partial y} \right\rVert_{H_1}\\+&\left\lVert \frac{\partial\hat{v}_{ij}}{\partial x} \right\rVert_{H_1}+\left\lVert \frac{\partial\hat{v}_{ij}}{\partial y} \right\rVert_{H_1}\Bigg)
    \end{aligned}
    \end{equation}
where $(\hat{u}_{ij}, \hat{v}_{ij})$ are the components of the predicted flow vector $\hat{\boldsymbol{y}}_{ij}$ at every pixel $j$.

Another common metric to quantify performance of optical flow algorithms \cite{middlebury} is the average angular error (AAE). The average angular error between two flow vectors is the average of the angle difference between every ground-truth and predicted flow vectors in the homogeneous coordinates, which are $\boldsymbol{y}_j^*=(u_j, v_j, 1)^T$ and $\hat{\boldsymbol{y}}_j^*=(\hat{u}_j, \hat{v}_j, 1)^T$ respectively. In Case II, the loss function $\mathcal{L}_3^i(Y_i, \hat{Y}_i)$ is defined as:
\begin{equation}
    \label{eq:average_angular_error}
    \begin{aligned}
        \mathcal{L}_3^i(Y_i, \hat{Y}_i) = \frac{1}{HW}\sum_{j=1}^{HW}\arctan{\left(\frac{\left\lVert\boldsymbol{y}_{ij}^*\times \hat{\boldsymbol{y}}_{ij}^*\right\rVert_2}{\boldsymbol{y}_{ij}^*\cdot\hat{\boldsymbol{y}}_{ij}^*}\right)}
    \end{aligned}
\end{equation}
The total loss function is then a weighted sum of the loss functions,
\begin{equation}
\label{eq:cyclic_aae}
\begin{aligned}
    J(X, \hat{X}, Y,\hat{\boldsymbol{Y}}) = \frac{1}{M}\sum_{i=0}^{M-1}\Big[\lambda_1\mathcal{L}_1^i(Y_i,\hat{\bm{Y}}_i) + &\lambda_2\mathcal{L}_2^i(X_{i+1}, \hat{X}_{i+1})\\ + &\lambda_3\mathcal{L}_3^i(Y_i, \hat{Y}_i)\Big]
    \end{aligned}
\end{equation}
We trained the network on only the faces dataset for 14 epochs and 228160 training pairs, with $\lambda_1=0.3$, $\lambda_2=0.5$, $\lambda_3=0.2$, and learning rate $2.5e{-6}$. We initialized the weights from the results of Experiment 2 (Case I), to see if there is any improvement in flow prediction after adding $\mathcal{L}_3$. 
In the next sections, we will use abbreviations for experiment and case numbers in the discussions for brevity. For example, Experiment 2, Case II is referred to as Exp. 2II, and no Roman numerals mean we refer to both cases of that particular experiment. 
\subsection{Micro-expression detection}\label{subsec:microexpression_method}
In this section, we describe how optical flow features are used for a micro-expression recognition task to demonstrate the efficacy of the optical flow generated using our method. The use of optical flow in micro-expression recognition has proven useful in several prior works, as described in section \ref{subsubsec:face_optical_flow_deep_learning}. 
\subsubsection{CNN and optical flow features}
\label{subsubsec:micro-expression_features}
To train the optical flow features, we use the three-dimensional lightweight CNN proposed by Liong \textit{et al}. \cite{optical_flow_cnn_3}, named the "Shallow Triple Stream Three-dimensional CNN", or STSTNet, which shows improved results compared with their previous work and other deep networks for micro-expression recognition. Their algorithm is evaluated on the CASME II \cite{CASME2}, SAMM \cite{SAMM}, and SMIC \cite{SMIC} datasets, composed of videos containing micro-expressions that represent either negative, positive, or surprise emotions (three-class classification). For each video sequence, they compute the optical flow between the onset the apex frames, and use this optical flow as input to train STSTNet classifier. The apex frames in SAMM are provided with the dataset, which is not the case with SMIC. The apex frames were also used for micro-expression recognition on SMIC dataset by Quang \textit{et al.} \cite{apex_frame_method}. We make use of their labeling for the SMIC dataset. We crop the faces based on keypoints obtained using the OpenFace 2.0 toolbox \cite{openface2} for SAMM. For SMIC, since OpenFace failed to detect the keypoints for some images, we instead use the dlib facial landmark detector \cite{dlib09}, which is based on an ensemble of regression trees \cite{kazemi}, and define the crop border at 15 pixels away from the maximum and minimum $x$ and $y$ image coordinates.

\begin{figure*}
 	\begin{minipage}[b]{1\linewidth}
 	\fbox{\includegraphics[keepaspectratio,width=1\textwidth,height=\dimexpr\textheight-2\baselineskip\relax]{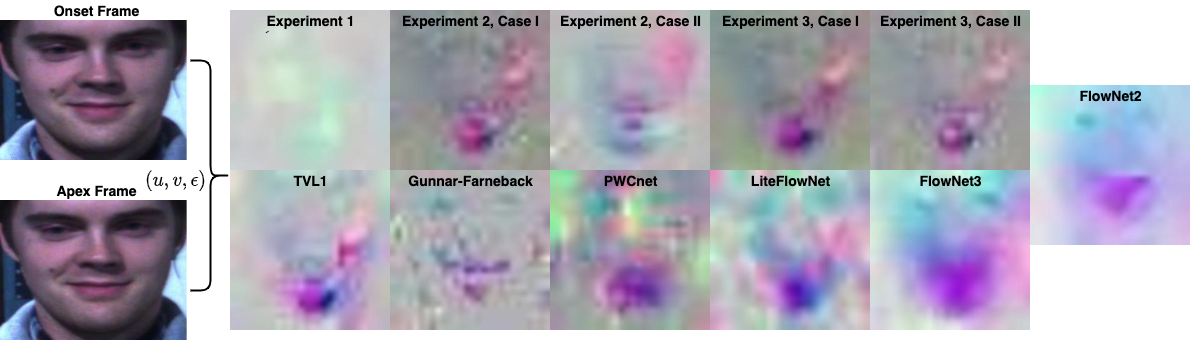}}
 	\end{minipage}
 	\caption{An example of the computed optical flow features used as inputs to train STSTNet for each network variant. Source: subject 03, SMIC \cite{SMIC}. \label{fig:ME_flow_examples}}
 \end{figure*}

We follow their recommended approach to train the STSTNet. The optical flow from the onset to the apex frame is used to compute the optical flow strain $\bm{\epsilon}(\bm{U})$ for a given flow field, $\bm{U}=(u(x,y), v(x,y))$. The strain is defined \cite{optical_flow_cnn_3} by the symmetric matrix known as the strain tensor
\begin{equation}
    \label{eq:optical_strain}
    \begin{aligned}
    \bm{\epsilon} &= \frac{1}{2}\left[\nabla\bm{U} + (\nabla\bm{U})^T\right]\\
    &= \begin{pmatrix}\frac{\partial u}{\partial x} & \frac{1}{2}\left(\frac{\partial u}{\partial y} + \frac{\partial v}{\partial x}\right) \\ \frac{1}{2}\left(\frac{\partial u}{\partial y} + \frac{\partial v}{\partial x}\right) & \frac{\partial v}{\partial y}\end{pmatrix}.
    \end{aligned}
\end{equation} The strain of a planar displacement field $(u, v)$ is well-known in solid mechanics, consisting of normal strains $\epsilon_{xx}, \epsilon_{yy}$, which are the diagonal elements, and shear strains $\epsilon_{xy}= \epsilon_{yx}$, which are the off-diagonal elements \cite{optical_flow_cnn_3}. The strain values represent the type of local deformation that occurs at each point in the flow field. The optical strain norm $\lvert\lvert\bm{\epsilon}(u(x, y), v(x, y))\rvert\rvert_s$ is then defined \cite{optical_flow_cnn_3} as:
\begin{equation}
    \lvert\lvert\bm{\epsilon}(u(x, y), v(x, y))\rvert\rvert_s = \sqrt{\epsilon_{xx}^2 + \epsilon_{yy}^2 + 2\epsilon_{xy}^2}
\end{equation}
The optical strain feature $\bm{V}\in\mathbb{R}^{H\times W \times 3}$ is an RGB image and, for a given pixel coordinate $(x_h, y_h)$, takes the value $(u(x_h, y_h), v(x_h, y_h), \lvert\lvert\bm{\epsilon}(u, v)\rvert\rvert_s)\in\mathbb{R}^3$. Fig. \ref{fig:ME_flow_examples} shows an example of the optical flow feature computed for an image pair, using the optical flow obtained from each network. In this example, the salient motion is an upwards curling of the lips plus a subtle leftwards shift in glance.

\subsubsection{Micro-expression detection experimental setup}
\label{subsubsec:microexpression_experiment}
The authors of STSTNet \cite{optical_flow_cnn_3} evaluate their model using leave-one-subject-out cross-validation (LOSOCV), and we do the same to train the micro-expression recognition networks. The SAMM and SMIC datasets were both used for the task. All optical flow networks described in Section \ref{subsec:experiment_2} are used separately to train STSTNet. For every optical flow network, we train the network three times: once on SAMM, once on SMIC, and once on the combined dataset consisting of both. We use the publicly available code provided by the authors \cite{optical_flow_cnn_3}, and thus replicate the exact same network architecture, with a learning rate of $5$e$-5$ and maximum epochs set to $500$. We note that the RGB input images, described in Section \ref{subsec:microexpression_method}, are resized to a resolution of $28\times28\times 3$. We also compute the TVL1 optical flow on SAMM and SMIC, as done in \cite{optical_flow_cnn_3}, to compare its performance with the optical flow features obtained from other networks. 


To deal with the class imbalance, we use macro-averaged recall, precision, and $F_1$-scores to evaluate the performance of every trained network. Additionally, the metrics specified by Yap \textit{et al.} \cite{microexpression_challenge} 
are the micro-averaged $F_1$-score and Unweighted Average Recall (UAR). The definition of UAR is equivalent to macro-averaged recall $R_M$. UAR is also popular for imbalanced multiclass problems, such as in \cite{optical_flow_cnn_3}, \cite{uar_1}, and \cite{uar_2}. 
The performance measures are defined as \cite{systematic_imbalance}:
\begin{equation}
\label{eq:macro_average_metrics}
\begin{aligned}
    &R_M = \frac{1}{n}\sum_{i=1}^n \frac{\sum_{j=1}^m tp_i^j}{\sum_{j=1}^m tp_i^j + fn_i^j},\;R_{\mu} = \frac{\sum_{i=1}^n \sum_{j=1}^m tp_i^j}{\sum_{i=1}^n \sum_{j=1}^m tp_i^j + fn_i^j},\\
    &P_M = \frac{1}{n}\sum_{i=1}^n \frac{\sum_{j=1}^m tp_i^j}{\sum_{j=1}^m tp_i^j + fp_i^j},\; P_{\mu} = \frac{\sum_{i=1}^n \sum_{j=1}^m tp_i^j}{\sum_{i=1}^n \sum_{j=1}^m tp_i^j + fp_i^j},\\
    &\;\;\;\;\;\;\;\;\;\;F_{1_x} = \frac{2P_xR_x}{P_x+R_x}, \;\;\;\;\; G_M = \sqrt[\leftroot{-1}\uproot{1}n]{\prod_{i=1}^n \frac{\sum_{j=1}^m t_{p_i}^j}{\sum_{j=1}^m t_{p_i}^j + f_{n_i}^j}}.
    \end{aligned}
\end{equation}
The subscripts $M,\, \mu$ denotes macro and micro-averaging, respectively, and for the $F_1$-score, $x\in\{M,\,\mu\}$. $tp_i^j$, $fp_i^j$, and $fn_i^j$ denote the true positive, false positive, and false negative of class $i$, sample $j$, for a total of $n\,(=3)$ classes and $m$ samples. Note that when the prediction for a given class is a true positive, this also counts as a true negative for each of the other two classes. The macro-averaged metrics tend to remove the bias caused by the imbalance degree, since it does not "favor" the classes with higher number of examples, as opposed to micro-averaging \cite{systematic_imbalance}. 

Moreover, since LOSOCV is used, this yields one metric per subject. We will combine the metrics to a single scalar, which we will refer to as the \textit{aggregated metric}. This has been done in other works such as \cite{losoc_application} (following a different experimental setup), and the aggregation is done by taking the mean of the metric across all subjects for every iteration.
\section{Results and discussion}\label{sec:ablation}
To evaluate the flow network, we first report preliminary results based on Experiment 1 and compare the flow predictions with networks trained on only the FlyingChairs or Sintel datasets. As indicated in section \ref{subsubsec:experiment_1}, the performance measure for this experiment is the average EPE. We then show the results of the ablation study for the experiments trained on the full dataset, and compare the networks using the average EPE and AAE. Next, we evaluate a number of other popular optical flow methods on the test set. These include FlowNet2.0 \cite{flownet2}, FlowNet3.0 \cite{flownet3}, LiteFlowNet \cite{liteflownet}, and PWC-Net \cite{optical_pyramid}, and the classic Gunnar-Farneback optical flow \cite{farneback}. Finally, the results of the micro-expression detection task using all networks are presented, which shows the usefulness of our method for a practical application.
\subsection{Results for ablation studies}\label{subsec:ablation_results}
We first describe the initial results of Exp. 1, which comprise of the network trained and tested on faces, FlyingChairs, and Sintel dataset. We report the performance in terms of average EPE. Table \ref{table:first_results} shows the first experiment's overall statistics for each network when tested on 3000 samples from our face dataset. We also compare the performance when trained and tested on other datasets. It is worth noting that the subjects that appear in the training set do not appear in the validation or test sets of our face data.

\begin{table}
\caption{The average EPE for each network described in subsection \ref{subsubsec:experiment_1}, trained and tested on all three datasets.\label{table:first_results}}
\centering
\begin{tabular}{ccccc}
    \hline
    &&&\multicolumn{1}{c}{\textbf{Tested on}}&\\
    \cline{3-5}
    &&Faces& \multicolumn{1}{c}{FlyingChairs} & \multicolumn{1}{c}{Sintel}\\
     \hline\hline
    \multicolumn{1}{c|}{}&\multicolumn{1}{c||}{Faces}& 0.4054&5.8495&5.1731\\
    \multicolumn{1}{c|}{\textbf{Trained on}}&\multicolumn{1}{c||}{FlyingChairs}&1.4040&1.4413&3.0300\\
    \multicolumn{1}{c|}{}&\multicolumn{1}{c||}{Sintel}&0.8282&7.7613&6.2358\\
    \hline
\end{tabular}
\end{table}
The error values in Table \ref{table:first_results} are in pixels, averaged over each of the test sets. Row 1 shows the results when the network was trained on faces and tested on all three datasets. Similarly, rows 2 and 3 are trained on FlyingChairs and Sintel and tested on all three. From Table \ref{table:first_results}, we observe that the network trained on our BP4D-derived face dataset performs best when tested on faces. This is likely due to the nature of the dataset the network was trained on. The flow fields on our face dataset consist of small, non-rigid motions, especially when the head motion is lacking, whilst the motion fields in the FlyingChairs dataset have larger magnitude and is more rigid. The Sintel dataset is also different in nature than the face dataset, but has smaller overall motion, and thus it is likely that the network trained on FlyingChairs overestimates the motion on the face dataset. Note that the results in Table \ref{table:first_results} are comparable to state of the art methods on the Sintel dataset, as can be seen in \cite{sintel}.

After adding the cyclic loss and training for more data and epochs, we expect to observe a difference in performance compared with Exp. 1. Here, we train the setup for Exp 1 again, using the same data split as the other experiments, for comparison purposes. Now we show the results of the networks trained with cyclic loss as described in sections \ref{subsubsec:experiment_2} and \ref{subsubsec:experiment_3}.

Table \ref{table:ablation_results} summarizes the statistics computed based on the results of Exp. 2 and Exp. 3. The statistics related to the flow fields (AAE and average EPE) are computed for all 32.5k image pairs in the test set.
\begin{table}
\caption{Flow performance for the ablation studies\label{table:ablation_results}}
\centering
\begin{tabular}{*3c}
\toprule
        Experiments&Ave. EPE& AAE\\
        \midrule
     Exp. 1 &0.2856 &0.1975 \\
     Exp. 2I&0.4610 &0.3033 \\
     \textbf{Exp. 2II} &\textbf{0.2498} &\textbf{0.1728} \\
     Exp. 3I &0.7010 &0.4524 \\
     Exp. 3II &0.4660 &0.2887 \\
     \bottomrule
\end{tabular}
\end{table}
As outlined at the end of section \ref{subsec:training_details}, Exp. 2I and Exp. 2II represent, respectively, the higher and lower reconstruction weight experiments, while Exp. 3I and Exp. 3II represent the experiment with smoothness constraint and the experiment with average angular error.

There are several observations to be made from these results. Adding the cyclic loss but with \textit{lower} reconstruction weights (Exp. 2II) improves the flow prediction compared to using only the EPE loss (Exp. 1), since both EPE and AAE decrease significantly. 
When there is higher weight on reconstruction loss (Exp. 2I), 
the network alters the predicted flow to improve 
the warped output's semblance to $X_2$. However, the higher focus on reconstruction worsens the performance of the AAE and EPE. One reason could be that the noisy ground truth does not necessarily reconstruct $X_2$ from $X_1$ very well, i.e. the reconstruction capability of a predicted flow field is adversarial to the ground-truth flow EPE and AAE.

Exp. 3 with the smoothness and AAE losses yields worse outcomes than the other two in terms of predicted flow, particularly compared to Exp. 2I. Note that Exp. 3 weights are initialized from the latter to test any change in performance. This could be due to the decreased weight in the EPE loss, which suggests that the EPE is a stronger indicator of flow performance than the AAE. The EPE encodes the direction in addition to the magnitude information. Another explanation would be that training data with angular error as a loss metric does not generalize well to test data, unlike 
the EPE. 
Exp. 3I exhibits the worst performance in both EPE and AAE amongst our network variants. This is likely due to the imposed smoothness constraints, which impose flow field values in the otherwise null regions outside the face boundary. 
\subsection{Comparison with other networks}\label{subsec:experiment_2}
We now compare the results with other notable optical flow implementations. Table \ref{table:comparison_results} shows the flow statistics computed for the network variants described earlier.\footnote{The interested reader is referred to the supplemental material, available at \url{https://www.dropbox.com/s/o7158gi46tppvb1/SupplementalMaterial_OpticalFlow.docx?dl=0}, for the error histograms for both ablation studies and comparison results.}
\begin{table}
\caption{Comparing various optical flow methods.\label{table:comparison_results}}
\centering
\begin{tabular}{*3c}
\toprule
        Optical flow methods&Ave. EPE& AAE\\
        \midrule
     \textbf{Exp. 2II (this work)} &\textbf{0.2498} &\textbf{0.1728} \\
     PWC-Net &1.1538 &0.4653 \\
     FlowNet2.0 &0.6719 & 0.4347\\
     FlowNet3.0-CSS &0.6839 &0.4457 \\
     LiteFlowNet &0.7226 &0.4771 \\
     Gunnar-Farneback &0.3670 &0.2294 \\
     \bottomrule
\end{tabular}
\end{table}
In all cases, the networks trained on our automatic face dataset perform better in both metrics than PWC-Net \cite{optical_pyramid} and LiteFlowNet \cite{liteflownet}, which are some of the popular CNN-based optical flow methods. PWC-Net demonstrates a notably high average EPE, but a more competitive AAE. This is likely due to an overestimation of the flow prediction magnitudes. FlowNet2.0 and FlowNet3.0-CSS, which are both state of the art improvements on FlowNetS, are both outperformed by all of our network variants with the exception of average EPE in Exp. 3I. The Gunnar-Farneback optical flow performs better than all methods in both average EPE and AAE, but is outperformed by Exp. 1 and Exp. 2I.
\begin{figure*}
 	\begin{minipage}[b]{1\linewidth}
 	\fbox{\includegraphics[trim=260 0 0 0,  clip,keepaspectratio,width=1\textwidth,height=\dimexpr\textheight-2\baselineskip\relax]{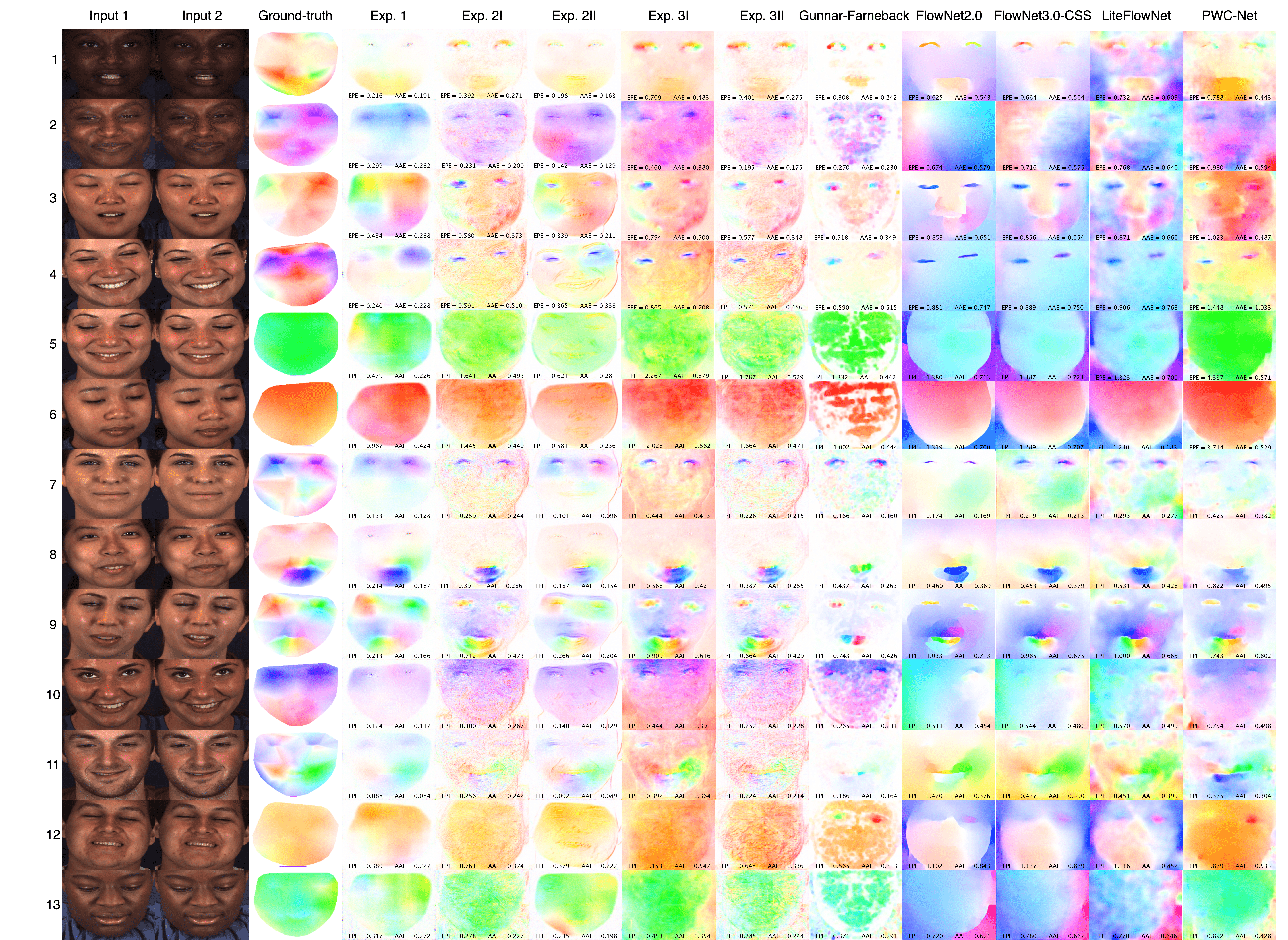}}
 	\end{minipage}
 	\caption{Color-coded optical flow predictions for a small subset of the test set for the networks trained in each of the experimental setups. The examples contain different types of facial motion, meant to illustrate the type of flow outputs produced by each network for qualitative assessment. \label{fig:color_coded_results}}
 \end{figure*}
 
To investigate the type of flow produced by each of the networks on the facial images, Fig. \ref{fig:color_coded_results} shows a sample subset of image pairs in the test set with their respective ground-truth and flow predictions from each network. The EPE and AAE for each prediction are also labeled. The saturation intensity in a given image is only representative of the intensity of that region relative to the other pixels of the same image. The same intensity in two images may have substantially different optical flow vector values. This is common practice in optical flow visualization, since it places emphasis on which motion is more salient for a given image. In images with small motion, as is the case in many frames in the BP4D dataset, using to-scale visualization would not convey important local motion information. We note that the following remarks for the remainder of this section are qualitative in nature and are based on a very small subset, but nevertheless yield some insight to accompany the statistics from Tables \ref{table:ablation_results} and \ref{table:comparison_results}.

We first observe the differences in flow predictions among the networks trained on our dataset. From these five, Exp. 1 shows the sparsest predictions, which is expected as it only minimizing the EPE from the sparse ground-truth flow. After introducing the cyclic loss in the other four experiments, denser optical features start to appear, caused by the added emphasis on image reconstruction. For example, this denser optical flow allowed the network to better predict the eye motion in rows 3 and 4 of Figure \ref{fig:color_coded_results}. Thus, the EPE loss taught the network to predict well the regional directions and magnitudes, and the cyclic loss helped it further localize the motion in these regions.

There is higher motion variance across the face in Exp. 2I compared to Exp. 2II. Although both were trained with cyclic loss, there is higher emphasis on the loss in Exp. 2I than in Exp. 2II, which more clearly shows the effect of the cyclic loss, since no other losses were introduced in Exp. 2. The outputs of the networks with cyclic loss also show coarser representations compared to the outputs of FlowNet2.0 and FlowNet3.0-CSS, such as in rows 9 and 10 in Fig. \ref{fig:color_coded_results}. When the smoothness constraints were imposed in Exp. 3I, the face segmentation learned by the network was affected, since the large values of the flow derivatives at the face boundaries enlargened the gradients in the smoothness loss function.

By both visual perception of these examples and the average EPE and AAE values from Table \ref{table:comparison_results}, the Gunnar-Farneback optical flow shows similarity in both direction and magnitude. Since the method is unbiased by any training data, this similarity provides a degree of validation to the ground-truth optical flow. However, it still underestimates optical flow in some instances, such as the near-zero regions in rows 6, 8, and 13. The Gunnar-Farneback flow also segments the face, since the background has zero motion. This is in contrast to the outputs of the other four networks (FlowNet2.0, FlowNet3.0-CSS, LiteFlowNet, and PWC-Net). The outputs of FlowNet2.0, FlowNet3.0-CSS, and LiteFlowNet all tend to estimate background flow. The flow trend of their outputs from the examples of Fig. \ref{fig:color_coded_results} can be matched with the outputs of the other networks, although some examples --- especially those with global motion, such as in rows 4, 5, and 13 --- show appreciable differences. PWC-Net demonstrates a more consistent flow pattern similar to our networks and Gunnar-Farneback. However, the EPE values in both the Fig. \ref{fig:color_coded_results} examples and Table \ref{table:comparison_results} suggest that the network, perhaps, overestimates the magnitude of the optical flow vectors in the field. Although its AAE is the second-highest, its value is close to several of the other methods. However, its EPE is significantly higher in comparison. This fact, complemented with the shown examples, suggests that the direction is a lesser problem than magnitude in PWC-Net.

The examples in rows 5, 6, 12, and 13 are characterized by predominantly global motion in one direction only. In these examples, the subject is mainly tilting their heads without any change in expression. Those in rows 4, 7, 8, 9 and 10 have the local motion as their salient feature, mainly in the eyes and mouth regions. Local face motion is more indicative of changes in facial expression, and the network's ability to identify the local motion can be used in FER. The remaining examples are rich with both global and local motions, indicating more aggressive motion along with the change in expressions. From these examples, all the networks were able to identify the local motions, except row 2, where three of the networks were not able to pick up the eye movements. The differences in network outputs are clearer in the examples with global motion.

From the overall results of the experiments, our networks trained on the automatically generated face dataset are better-suited at predicting the optical flow on faces compared to other networks. 
\subsection{Results of micro-expression detection}
\label{subsec:microexpression_results}
We now report on the results of the aforementioned experiments for micro-expression detection. The micro- and macro-averaged metrics are shown for every network on each of the SAMM, SMIC, and combined datasets in Tables \ref{table:microexpression_results_separate} and \ref{table:microexpression_results_combined}. In these tables, the aggregation of the metrics across the subjects from the LOSOCV is the mean of the metric across the subjects.
\begin{table*}[t]
\caption{Results of the aggregated performance metrics for micro-expression recognition on the \textbf{SAMM} and \textbf{SMIC} datasets separately, using TVL1 optical flow as done in \cite{optical_flow_cnn_3}, our network with different variants, and the other optical flow CNN architectures.\label{table:microexpression_results_separate}}
\centering
\begin{tabular}{ccccccccccc}
\toprule
&&\multicolumn{3}{c}{SAMM}&\multicolumn{1}{c|}{}&&\multicolumn{3}{c}{SMIC}&\\
    Optical Flow Methods& $P_M$& $R_M$ & $F_{1_M}$& $F_{1_{\mu}}$&\multicolumn{1}{c|}{$G_M$}&$P_M$& $R_M$ & $F_{1_M}$& $F_{1_{\mu}}$&$G_M$\\
        \midrule
     TVL1 &0.800 &\textbf{0.777} & 0.773 & 0.698 & \multicolumn{1}{c|}{0.513} &0.612&0.606&0.589&0.549&0.385 \\
     Exp. 1 & \textbf{0.888} & 0.737 & \textbf{0.794}  & \textbf{0.727} &  \multicolumn{1}{c|}{0.429} &0.480&0.518&0.489&0.404&0.263 \\
     Exp. 2I & 0.763 & 0.713 & 0.722 & 0.636 & \multicolumn{1}{c|}{0.411}&0.610&0.585&0.581&0.518&0.395  \\
     Exp. 2II &\textbf{0.849} &0.745 & \textbf{0.784} & \textbf{0.713} & \multicolumn{1}{c|}{0.487} &0.582&0.543&0.546&0.432&0.280\\
     Exp. 3I & 0.765 & 0.735&0.738 & 0.653& \multicolumn{1}{c|}{0.457} &0.621&0.591&0.593&0.538&0.350\\
     Exp. 3II & \textbf{0.816} &\textbf{0.770} & \textbf{0.780}  &\textbf{0.715}  &\multicolumn{1}{c|}{\textbf{0.516}} &0.603&0.580&0.577&0.514&0.364  \\
     Gunnar-Farneback & 0.719& 0.696 & 0.690& 0.608 & \multicolumn{1}{c|}{0.412} &0.570&0.479&0.505&0.387&0.217\\
     FlowNet2.0 &0.724 & 0.717 & 0.695 & 0.650 &\multicolumn{1}{c|}{0.431} &\textbf{0.641}&\textbf{0.627}&\textbf{0.622}&\textbf{0.552}&\textbf{0.420} \\
     FlowNet3.0-CSS &0.813 &\textbf{0.764} &0.773 &0.711& \multicolumn{1}{c|}{\textbf{0.528}} &\textbf{0.633}&\textbf{0.659}&\textbf{0.636}&\textbf{0.592}&\textbf{0.493} \\
     LiteFlowNet &0.767&0.761&0.749&0.694&\multicolumn{1}{c|}{\textbf{0.514}} &\textbf{0.635}&\textbf{0.658}&\textbf{0.637}&\textbf{0.607}&\textbf{0.438}\\
     PWC-Net &0.742&0.738&0.722&0.676&\multicolumn{1}{c|}{0.485} &0.501&0.525&0.501&0.422&0.249\\
     \bottomrule
\end{tabular}
\end{table*}
\begin{table}[t]
\caption{Results of the aggregated performance metrics for micro-expression recognition on the \textbf{combined SAMM and SMIC dataset}, using TVL1 optical flow as done in \cite{optical_flow_cnn_3}, our network with different variants, and the other optical flow CNN architectures.\label{table:microexpression_results_combined}}
\centering
\resizebox{1\textwidth}{!}{\begin{minipage}{\textwidth}
\resizebox{0.5\columnwidth}{!}
{\begin{tabular}{*6c}
\toprule
&&\multicolumn{3}{c}{SAMM and SMIC}&\\
        Optical Flow Methods& $P_M$& $R_M$ & $F_{1_M}$& $F_{1_{\mu}}$&$G_M$\\
        \midrule
     TVL1 &\textbf{0.740}&\textbf{0.714}&\textbf{0.711}&\textbf{0.662}&\textbf{0.491} \\
     Exp. 1 &0.692&0.601 &0.622  &0.548  &0.307   \\
     Exp. 2I &0.726 &0.700 & 0.694 & 0.635 & 0.465 \\
     Exp. 2II &\textbf{0.757}&0.674&0.702&0.614&0.342 \\
          Exp. 3I  &0.726  &0.706  &0.702  &\textbf{0.661} &0.459\\
     Exp. 3II &0.704 &0.678 &0.676  &0.592   &0.447  \\
     Gunnar-Farneback &0.720&0.700&0.690&0.629&0.472 \\
     FlowNet2.0 &\textbf{0.745}&0.709&\textbf{0.709}&0.636&0.459 \\
     FlowNet3.0-CSS &0.725&\textbf{0.727}&\textbf{0.713}&\textbf{0.665} &\textbf{0.511}\\
     LiteFlowNet &0.717&\textbf{0.717}&0.702&0.650&0.446 \\
     PWC-Net &0.704&0.693&0.687&0.631&\textbf{0.482} \\
     \bottomrule
\end{tabular}}
\end{minipage}}
\end{table}
The results in Table \ref{table:microexpression_results_separate} indicate that the performance of STSTNet trained on optical flow features from different networks also significantly depends on the dataset it is trained on. For each evaluation measure, the top three performing networks are indicated in bold. We note that the $F_1$-scores are typically lower than the precision and recall since these are aggregated metrics, i.e. the $F_1$-score averaged over all $F_1$-scores in the LOSOCV, and is not the harmonic mean of the aggregated precision and recall. 

For macro-averaged precision $P_M$, as well as the macro and micro-averaged $F_1$-scores on the SAMM dataset, the top scores are achieved from Experiments 1, 2II, and 3II, followed closely by FlowNet3.0-CSS and TVL1. The higher $F_1$-scores are more influenced by the precision values and less so by the recalls. 
Exp. 1 is the highest for these three metrics, while TVL1 scored highest in $R_M$, and FlowNet3.0-CSS in geometric mean. This is one testimony to the complexity of capturing the overall classification performance with a single scalar metric for multi-class problems, since the proposed metrics can each emphasize different features of the classifier performances. Exp. 3II is the only variant which is consistently among the top 3 for all metrics, at either second or third.

The SMIC results allow for a more consistent inference on the performance of the classifiers. Across all metrics, the top three networks were FlowNet2.0, FlowNet3.0-CSS, and LiteFlowNet. Both the precision and recall, and consequently the $F_1$-scores, follow more similar trends, in contrast with the SAMM and combined training protocols. For precision, recall, and $F_1$-scores, the lowest three scores are interchanged amongst Exp. 1, PWC-Net, and Gunnar-Farneback. In fact, Gunnar-Farneback and PWC-Net are consistently the least performing across all three training protocols.

By comparing the results across the three training protocols, it is difficult to conclude that optical flow features computed from one specific method will be optimal for training the STSTNet classifier for micro-expression detection. Although the networks trained using our method performed well when trained and tested on SAMM, they were somewhat outperformed in the other two protocols. However, even in these cases, they were not as consistently behind when compared to Gunnar-Farneback and PWC-Net. This could be due to the sparse nature of the learned optical flow representations from our generated dataset. It is also plausible that the accuracy of the flow magnitude prediction is not a consistent predictor of its performance on micro-expression detection. We hypothesize that our method will overcome the performance difference in some of the results if we use a denser keypoint tracker during the optical flow training phase to generate the BP4D ground-truth. This will likely improve the network's ability to more consistently capture fine local facial motion which may otherwise have been missed in the current work. Furthermore, as previously discussed, we have used FlowNetS to train the face data to benchmark its efficacy compared to other networks, and thus using a better-designed CNN along with the denser keypoint ground-truth will likely further improve the performance. 
\section{Conclusion and Future Work}\label{sec:conclusion}
In this paper, we explore the possibility of using a facial expression dataset to learn optical flow representations based on a self-supervised technique. Motion information on faces has been shown to be useful in facial expression analysis in multi-modal techniques.

The dataset is generated by using the image sequences from the BP4D-Spontaneous dataset to compute the optical flow ground-truth. The OpenFace 2.0 toolbox, which uses a constrained local model, is used to locate the facial landmarks on every image. Delaunay triangulation is then used on the resulting set of points to form the face mesh and allow the computation of the optical flow for every pair of images using triangle-to-triangle affine maps to develop an automatic facial optical flow dataset. The generated dataset, with a total of nearly 324k image pairs, is used as a \emph{noisy} ground-truth for optical flow to train the FlowNetS convolutional autoencoder architecture with 228k pairs in the training partition. 

It was observed that training the FlowNetS architecture for optical flow on this automatically generated noisy ground-truth data improved the network's ability to predict optical flow on face data in particular. The learned representations also helped the network give good accuracy on the FlyingChairs and Sintel datasets. This demonstrates that the facial movements are nicely encoded in our data which enables the network to learn subtle movements that are useful on the challenging Sintel dataset as well. A cyclic loss was also added for optimization to help the network use the predicted flow to reconstruct the second image, and the flow results from different experimental setups are compared. It was seen that the flow predictions are best when there is less emphasis on reconstruction, due to denser representations learned with reconstruction that are not present in the ground-truth flow fields. Compared with other optical flow methods (Gunnar-Farneback, FlowNet2.0, FlowNet3.0-CSS, LiteFlowNet, and PWC-Net), it was shown that the networks trained on the generated dataset predict better flow representations, as quantified by the flow error metrics. This implies that a network trained on good face optical flow ground-truth have the propensity to outperform networks trained on other datasets. 

To investigate the performance of the different optical flow network variants in an FER application, the optical flow features were used to train STSTNet for micro-expression detection. The experimental results using different performance metrics were mixed, e.g. FlowNet3.0-CSS outperformed the other methods in a good proportion of the cases. However, our method also demonstrated promising results in some cases, and note that further improvements and extension to this baseline work can help improve its application to FER.

For further investigation and improvement, future work related to this work can include the following:
\begin{enumerate}
    \item Use a denser tracker such as Zface \cite{zface} to track a higher number of key-points for a finer triangulation and denser optical flow ground-truth in our automatic data generation algorithm.
    \item Use a more complex CNN architecture to train the denser optical flow ground-truth.
    \item Train the optical flow network on faces with some head rotation, such as pan and tilt, to learn optical flow for non-frontal faces.
    \item Tackle challenges in optical flow learning, such as in environments with occlusion and illumination, to increase the robustness of facial optical flow.
\end{enumerate}

In addition to these improvements for optical flow learning, the empirical analysis can be extended to evaluate the performance of the face-trained optical flow CNN in other problems in facial expression analysis, such as action unit recognition.

%


\ifCLASSOPTIONcompsoc
  \section*{Acknowledgments}
\else
  \section*{Acknowledgment}
\fi

This work was supported in part by the American University of Sharjah, FRG17-R44 research grant.

\ifCLASSOPTIONcaptionsoff
  \newpage
\fi



%
\bibliographystyle{IEEEtran}
\bibliography{IEEEabrv,ref}


%

\begin{IEEEbiography}[{\includegraphics[width=1in,height=1.25in,clip,keepaspectratio]{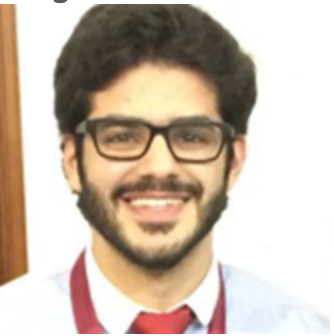}}, \vspace{5mm}]{Muhannad Alkaddour}
received the M.S. degree from the Mechatronics Engineering Graduate Program at the American University of Sharjah (AUS) in 2020. His research experience and interests are in artificial intelligence, with emphasis on deep learning and computer vision, as well as robotics, control systems, and mechanical vibrations.
\end{IEEEbiography}
\begin{IEEEbiography}[{\includegraphics[width=1in,height=1.25in,clip,keepaspectratio]{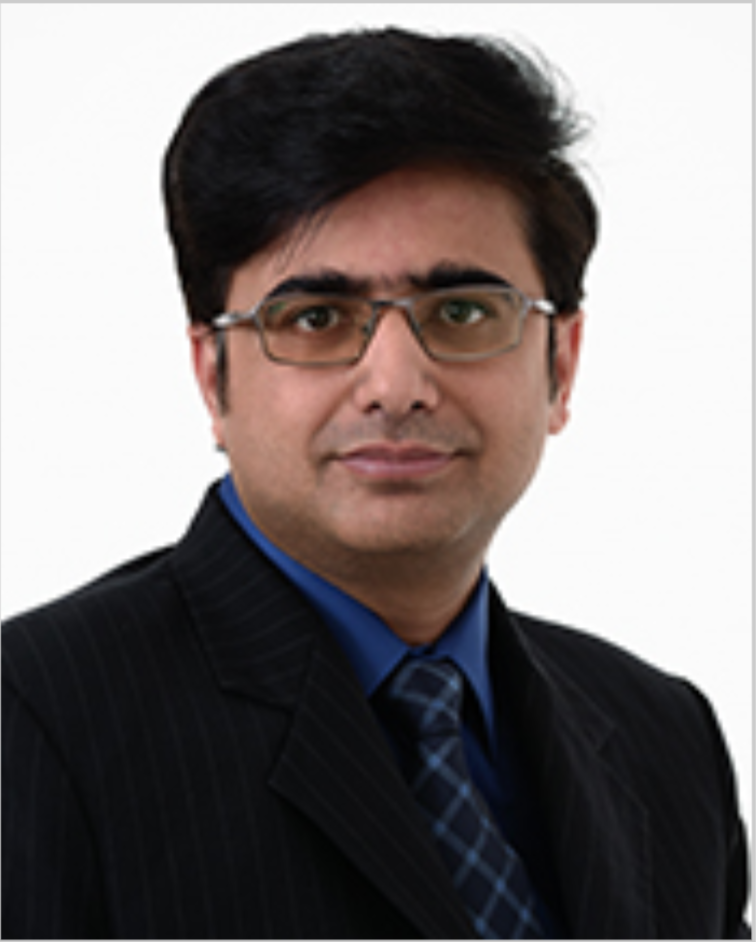}}]{Usman Tariq}
received the M.S. and Ph.D. degrees from the Electrical and Computer Engineering Department, University of Illinois at Urbana-Champaign (UIUC), in 2009 and 2013, respectively. He is currently a Faculty Member of
the Department of Electrical Engineering, American University of Sharjah (AUS), UAE. Before AUS, he worked as a Research Scientist with
Computer Vision Group, Xerox Research Center Europe, France. His research interests include computer vision, image processing, and machine learning, in general; while facial expression recognition and face biometrics, in particular.
\end{IEEEbiography}
\begin{IEEEbiography}[{\includegraphics[width=1in,height=1.25in,clip,keepaspectratio]{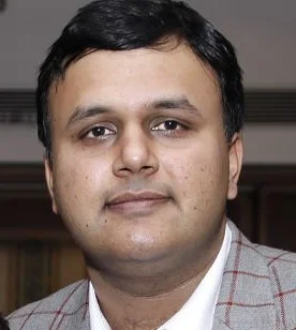}}]{Abhinav Dhall}
received the PhD degree in computer science from the Australian National University, Canberra, Australia, in 2014. He is currently a postdoctoral research fellow at the Vision and Sensing Group, Human-Centred Technology Research Centre, University of Canberra, Bruce, Australia, and an adjunct
research fellow at the Australian National University. He was awarded the Best Doctoral Paper Award at ACM International Conference on Multimodal Interaction 2013, Best Student Paper Honourable mention at IEEE International Conference on Automatic Face and Gesture Recognition 2013 and Best Paper Nomination at IEEE International Conference on Multimedia and Expo 2012. His research interests are in computer vision for affective computing and social signal processing. He is a member of the IEEE.
\end{IEEEbiography}




\end{document}